# Construction of unbiased dental template and parametric dental model for precision digital dentistry

Lei Ma[a,1], Jingyang Zhang[a,1], Ke Deng[b], Peng Xue[a], Zhiming Cui[a], Yu Fang[a], Minhui Tang[a], Yue Zhao[c], Min Zhu[d], Zhongxiang Ding[e], Dinggang Shen[a,f,g,*]

[a]*School of Biomedical Engineering, ShanghaiTech University, Shanghai 201210, China.*
[b]*Division of Periodontology and Implant Dentistry, Faculty of Dentistry, The University of Hong Kong, Hong Kong SAR 999077, China.*
[c]*School of Communication and Information Engineering, Chongqing University of Posts and Telecommunications, Nan'an District, Chongqing, 400065, China.*
[d]*Shanghai Ninth People's Hospital, Shanghai Jiao Tong University, Huangpu District, Shanghai, 200011, China.*
[e]*Department of Radiology, Hangzhou First People's Hospital, Zhejiang University, Hangzhou, 310006, China.*
[f]*Shanghai United Imaging Intelligence Co., Ltd., Shanghai 200230, China, and Shanghai Clinical Research and Trial Center, Shanghai 201210, China.*
[g]*Shanghai Clinical Research and Trial Center, Shanghai, 201210, China.*

**Abstract**

Dental template and parametric dental models are important tools for various applications in digital dentistry. However, constructing an unbiased dental template and accurate parametric dental models remains a challenging task due to the complex anatomical and morphological dental structures and also low volume ratio of the teeth. In this study, we develop an unbiased dental template by constructing an accurate dental atlas from CBCT images with guidance of teeth segmentation. First, to address the challenges, we propose to enhance the CBCT images and their segmentation images, including image cropping, image masking and segmentation intensity reassigning. Then, we further use the segmentation images to perform co-registration with the CBCT images to generate an accurate dental atlas, from which an unbiased dental template can be generated. By leveraging the unbiased dental template, we construct parametric dental models by estimating point-to-point correspondences between the dental models and employing Principal Component Analysis to determine shape subspaces of the parametric dental models. A total of 159 CBCT images of real subjects are collected to perform the constructions. Experimental results demonstrate effectiveness of our proposed method in constructing unbiased dental template and parametric dental model. The developed dental template and parametric dental models are available at https://github.com/Marvin0724/Teeth_template.

## 1. Introduction

Dental diseases, such as malocclusion or edentulism, affect a large number of people worldwide, leading many to seek dental treatments to restore oral functions and improve facial aesthetics [1]. Digital dentistry is an emerging technology reforming today's dental treatments because of its efficiency and accuracy [2]. Typical digital dentistry process involves three main steps: digital dental image acquisition, dental image processing and analysis, and transfer of digital treatment to the clinical environment [3]. By leveraging digital imaging techniques, dental professionals can improve accuracy and efficiency of their diagnosis, treatment planning, and treatment deliv-ery, ultimately leading to better patient outcomes [4].

*Corresponding author.
Email address:* Dinggang.Shen@gmail.com (Dinggang Shen)
[1] Equal contribution

Dental template is a standardized, three-dimensional (3D) representation of a typical dental model. It usually serves as a valuable tool for spatial normalization and spatial correspondence estimation across different subjects for dental disease diagnosis and treatment planning in many dental applications, such as dental implant planning and orthodontic treatment planning [5]. However, currently, there is still no dental template that can represent population-average dental structures without subject-biases, i.e., unbiased dental template. This is because it is still challenging to construct an unbiased dental template despite advances in dental imaging technologies, as explained below. First, the maxilla and mandible that support the upper and lower rows of teeth, have complicated anatomical and morphological structures. Second, the volume ratio of teeth with respect to the whole image is small. These two factors make it challenging to accurately construct the unbiased dental template. Previously, researchers were limited to using biased dental templates, such as artificial templates or selected subject templates, which could potentially influence the accuracy of the final results. For example, Barone et al. created a digital dental template from a reference physical model to reconstruct complete 3D teeth model from panoramic radiographs [6]. Wu et al. artistically designed templates for four categories of teeth (i.e., incisors, canines, premolars and molars), and used these templates to fit the teeth model in the plaster cast scans [7]. Pei et al. created a 3D dental template by digitizing the teeth specimens used in dental clinics, and applied it to constrain tooth segmentation from CBCT images [8]. Gholamalizadeh et al. randomly selected a typical dental model as the reference dental template for spatial alignment [5]. To learn the tooth arrangement in digital orthodontics, Wei et al. normalized the position and orientation of different dental models by defining a local coordinate system on each model and aligning it with the world coordinate system [9]. An unbiased dental template would improve the normalization of dental models, thus resulting in higher prediction accuracy of deep learning methods [10].

Parametric dental models, which describe shape and position variations of dental models in a population, also have broad applications in digital dentistry, such as teeth segmentation [11], dental mesh repairing [12] and 3D teeth reconstruction [13, 7]. The construction of parametric dental models involves the following steps [14, 15]: (1) rigidly aligning all the dental models to a common space, (2) estimating the point-to-point correspondence between the aligned dental models and (3) employing Principal Component Analysis (PCA) [16] to determine subspaces of the parametric dental models. However, the accuracy of existing parametric dental models is limited due to the lack of an unbiased dental template, which is essential for precise spatial alignment and point-to-point correspondence estimation. Without an unbiased template, the construction of parametric dental models becomes challenging, and the resulting models may contain subject-specific biases. Therefore, the development of an unbiased dental template is crucial for construction of accurate parametric dental models and for improving dental image processing and analysis.

Anatomical atlases, i.e., population-average anatomical structures, have been widely used as unbiased templates for medical image processing and analysis [17]. These templates provide a population-average representation of typical shapes and locations of a specific anatomical structure, while maintaining high cross-individual validity within the population [18]. Therefore, the unbiased template is an effective tool for medical image processing, like spatial normalization and spatial correspondence estimation across different subjects. Over the years, numerous unbiased templates of different anatomies have been constructed for various medical applications. For instance, Rajashekar et al. constructed FLAIR MRI and noncontrast computed tomography (CT) atlases of elderly populations for clinical applications [17]. Chen et al. developed a 4D infant brain atlas for dynamic brain development analysis during infancy [18]. Additionally, Lee et al. created a CT atlas of cardiac disease understanding and abdominal atlas for integration of tissue information in different scales [19].

In this paper, we develop an unbiased dental template by constructing an accurate dental atlas from CBCT images. To address the challenges of complex adjacent anatomical structures and low volume ratio, we propose to enhance CBCT images and their teeth segmentation images through image cropping, image masking and segmentation intensity reassigning. Then, we construct a dental atlas of CBCT images by performing registration between the enhanced segmentation images to guide the registration between the masked CBCT images, which further reduces influence of the adjacent anatomical structures on atlas construction and improves accuracy of teeth region registration. Leveraging the constructed unbiased dental template, we build parametric dental models for dentition and each individual tooth. The contributions of this paper can be concluded as follows: (1) We develop an unbiased dental template by accurately constructing a dental atlas from CBCT images. (2) We propose an approach to enhance CBCT images and teeth segmentation images for accurate dental atlas construction. (3) By using the unbiased dental template, we construct parametric dental models by analyzing the shape variations of dentition and every tooth. To the best of our knowledge, this is the first study for unbiased dental template construction.

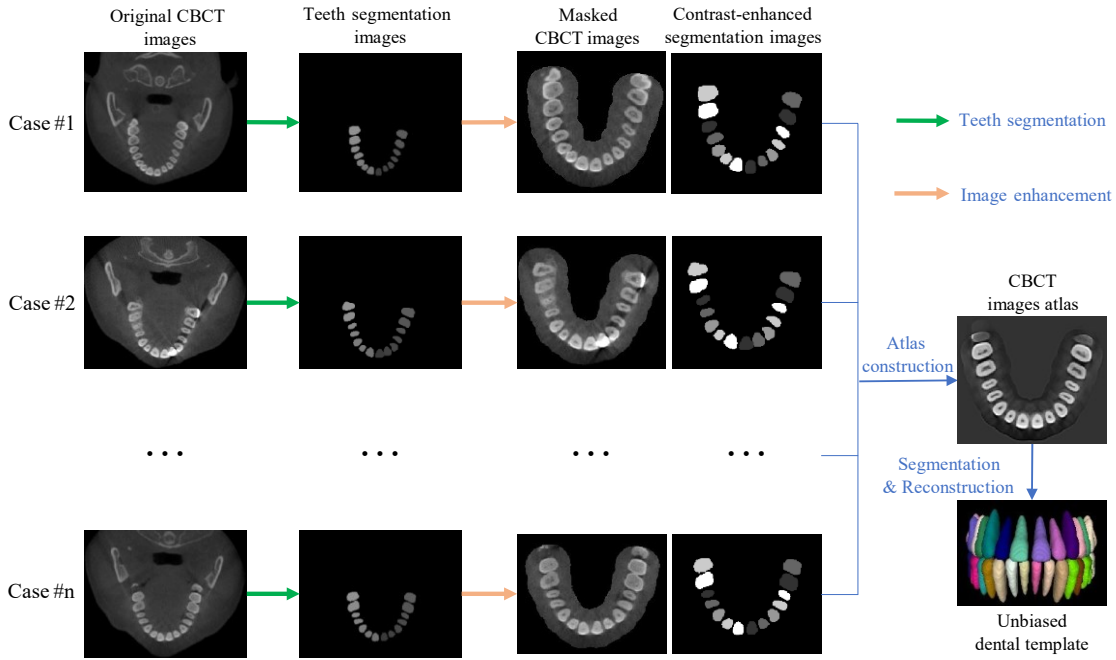

Figure 1: The proposed framework for unbiased dental template construction. The image enhancement includes image cropping, image masking and segmentation intensity reassignment. Image intensity values of all images shown in this figure were remapped to the full display range.

The rest of the paper is organized as follows. The studied data and proposed method are described in Section 2. Section 3 presents the results of dental template and parametric model construction. The paper is finally discussed and concluded in Section 4.

## 2. Materials and Method

### 2.1. Dataset

In this study, we selected cone-beam computed tomography (CBCT) images of 159 subjects from our digital archive. All the CBCT images were acquired by a dental CBCT scanner (Manufacturer: Planmeca) in First People's Hospital of Hangzhou. During the selection, we excluded the subjects with severe malocclusion or with missing of 2 more teeth. Additionally, we did not consider the third molars in this study. The age range of the selected subjects was 18 to 60, and all the subjects were of Chinese ethnicity. The resolution of the CBCT images varied from $0.1mm$ to $0.4mm$ with X and Y dimensions both set to 400, while the Z dimension ranged from 256 to 328. Prior to the dental atlas construction, we pre-processed the CBCT images by first standardizing their resolutions to $0.4mm \times 0.4mm \times 0.4mm$ and then normalizing voxel intensities of the images to the range of [0,1].

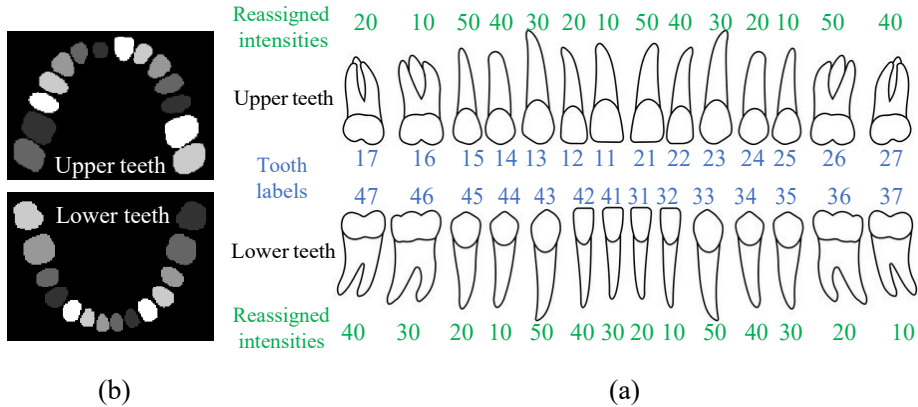

Figure 2: Intensity reassignment of teeth segmentation images. (a): An example of a paired upper and lower teeth slices images with reassigned intensities. (b): teeth numbering and their corresponding intensity reassignment. Note that image intensity values are remapped to the full display range.

## 2.2. Dental Atlas Construction

Figure 1 illustrates the framework proposed for constructing a dental atlas. First, we employ a deep learning-based tooth segmentation method [20] to segment teeth from the CBCT images. Then, based on the teeth segmentation, we enhance CBCT images and their segmentation images for accurate atlas construction, which includes cropping the images, masking the surrounding tissues, and enhancing the intensity contrast between adjacent teeth in the segmentation images. Next, we utilize the SyGN atlas construction algorithm [21] to perform co-registration between the CBCT and segmentation images, resulting in a CBCT image atlas and a segmentation image atlas. Finally, we manually segment teeth from CBCT image atlas to obtain an unbiased dental template.

### 2.2.1. Teeth Segmentation and Labeling:

In this study, we propose to use teeth segmentation to guide image enhancement and co-registration for accurate dental atlas construction. For this purpose, we adopt a fully automatic deep learning-based tooth segmentation method to segment all the teeth from the normalized CBCT images [20]. The segmentation method employs an U-Net network to extract the teeth region and uses a hierarchical morphology-guided network to differentiate and segment each tooth from the extracted teeth region. This approach enables accurate segmentation of all teeth, which is essential for creating an accurate dental atlas.

In order to identify the segmented teeth, we deploy a hierarchical method to iteratively label the teeth in the segmentation images. First, we randomly select 30 subjects and manually label the segmented teeth in those subjects. Using the labeled segmentation images, we create an initial dental atlas of teeth segmentation images. Then, we apply atlas-based labeling to the remaining segmentation images by non-rigidly registering the initial dental atlas to those images [22]. Finally, we manually correct wrong labels during a final quality checking of the results.

### 2.2.2. Segmentation-Guided Image Enhancement:

Prior to constructing dental atlas, we perform segmentation-guided image processing to enhance CBCT and segmentation images for accurate atlas construction. First, we crop both CBCT and segmentation images with a margin of 30 in the x, y, and z directions around the segmented teeth, thus removing the background region. Next, we eliminate the tissue surrounding the teeth by using the teeth segmentation images as masking images to mask the cropped CBCT images. Here, we dilate the binary images of the teeth segmentation before performing the masking operation. The purpose of the dilation operation is to avoid masking the teeth region that are not accurately segmented.

To further improve accuracy of the co-registration between CBCT and teeth segmentation images, we enhance intensity contrast between one tooth and its adjacent teeth in the segmentation images by reassigning intensity val-ues to the teeth according to their labels. Fig. 2(a) shows a teeth segmentation image with reassigned intensity values using a tooth label-based assignment strategy (Fig. 2(b)). High intensity contrast between adjacent teeth can push teeth in the moving image to their corresponding teeth in the fixed image during the registration, thus improving the accuracy of dental atlas construction.

*2.2.3. Atlas Construction*

The enhanced CBCT images and teeth segmentation images are used to jointly construct a teeth segmentation image atlas and a CBCT image atlas. The symmetric group-wise normalization (SyGN) based method in ANTs toolkit is adopted to construct the dental atlases with unbiased shape and appearance [21]. The SyGN-based method iteratively performs multi-channel registration between the CBCT images and the teeth segmentation images, and involves multiple iterations of registration, averaging, and optimization with the following steps:

(1) Initial templates are first generated for the CBCT and teeth segmentation images by averaging all the images in their respective domain.

(2) Each pair of the CBCT and segmentation image is aligned to their respective initial templates using the symmetric normalization (SyN) algorithm [23]. The alignment includes rigid, affine, and non-rigid diffeomorphic registrations.

(3) The warped CBCT images and segmentation images are averaged to generate a new segmentation template and a new CBCT image template.

(4) The affine transformations and deformation fields obtained in (2) are averaged and then applied to the templates achieved in (3) for template optimization.

(5) The CBCT and segmentation templates outputted at (4) are used as initial templates for next iteration.

The aforementioned operations are iterated to generate an accurate CBCT image atlas. Cross correlation is selected as the similarity metric for the image registration. The shrinkage factors, smoothing factors and max iterations [21] of the SyN method are set to 8×4×2×1, 3×2×1×0 and 100×80×40×10, respectively. Finally, we segment teeth from the outputted CBCT image atlas and then reconstruct tooth mesh models from the segmentation image, resulting in an unbiased dental template.

*2.3. Parametric Dental Model*

Human adults typically have 28 teeth if the third molars are not considered. They can be divided into four categories, i.e., incisors, canines, premolars and molars. By leveraging the constructed unbiased dental template, we are able to build precise parametric dental models, which includes a parametric dentition model and a parametric model for each tooth, from real clinical dataset. The dentition parametric model encodes the pose and shape variation of all teeth in the upper and lower rows. The parametric tooth model describes the shape variations of one tooth among the population.

Figure 3 illustrates the framework proposed for constructing parametric dental models, i.e., parametric dentition model and parametric tooth model. To build the parametric models, we need to estimate point-wise correspondences between the teeth of different subjects in the dataset. For this purpose, we first reconstruct dental models from the teeth segmentation images, and then rigidly align the dental models to the constructed unbiased dental template. To construct the parametric dentition models, we non-rigidly register each tooth in the constructed dental template to its corresponding tooth in the aligned teeth models through coherent point drift (CPD) algorithm [24], which results in dental models with strict point-to-point correspondence. Once the correspondence is established, PCA is employed to determine the shape variation modes of the dentition models as the subspaces of their parametric models. To construct the parametric tooth models, we rigidly register each tooth in the aligned dental models to its corresponding tooth in the teeth template, and then estimate point-to-point correspondences between the registered teeth for PCA analysis.

## 3. Results

*3.1. Unbiased Dental Template*

We followed the proposed dental atlas construction framework to create a dental atlas of CBCT image using the studied dataset. The radius of the structuring element in the dilation operation was set to 15. The itera-tions of the SyGN-based method was set to 10. According to the results, the final CBCT image atlas had a size of 240×215×172 and a resolution of 1×1×1 $mm^3$. To reconstruct an unbiased dental template from the dental atlas, we performed a semi-automatic teeth segmentation process. We first applied the learning-based method [20] to segment teeth automatically, and then refined the segmentation result manually.

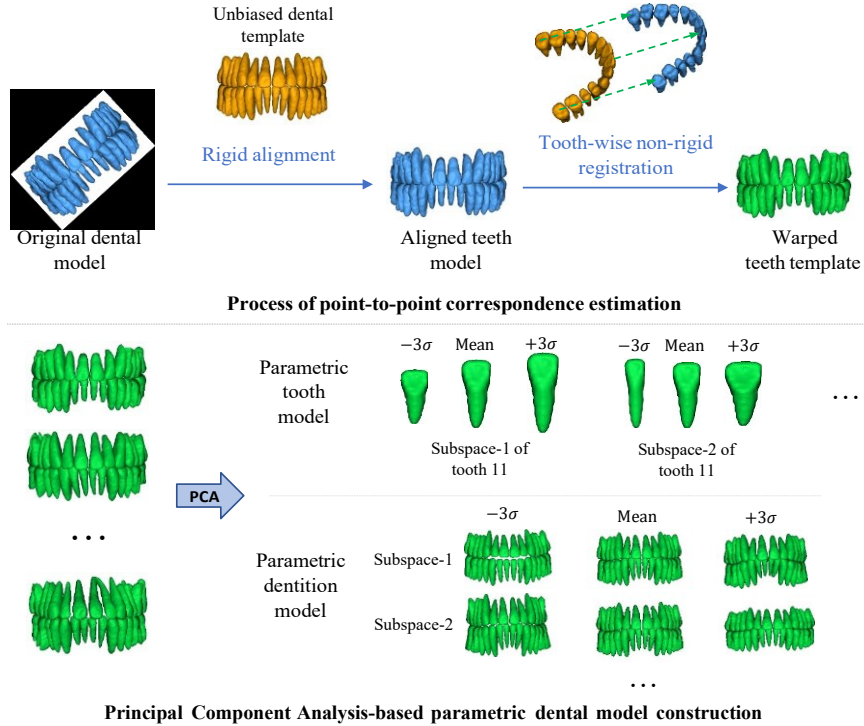

Figure 3: The framework of parametric dental model construction. The framework consists of two steps, i.e., point-to-point correspondence estimation and Principal Component Analysis-based subspaces determination.

The constructed dental atlas is presented in Fig. 4. The first row displays slice images of a selected CBCT image template, while the second and third rows show the slice images of the constructed CBCT image atlas and their corresponding teeth segmentation (slice) images, respectively. According to the results, we can observe that the constructed CBCT image atlas provides clearer anatomical structures with distinct boundaries compared to the selected CBCT images. Moreover, the teeth in the dental atlas are arranged in a more smooth and aesthetic dental arch curve [25].

Fig. 5 displays the unbiased dental template derived from the CBCT atlas image. The first row shows the different views (i.e., front view, back view, side view and bottom view) of the dental template. The dental template showcases an ideal arrangement of teeth, with no overlapping or malpositioning. Specifically, the front and back views of the dental template reveal the upper front teeth slightly overlapping the lower ones, and the upper and lower dental midlines aligning with each other. From the side view, all upper teeth perfectly fit the upper ones and the teeth are aligned with a nice curve. The second and third rows of Fig. 5 provide detailed views of the individual tooth templates in the upper and lower rows, respectively.

We conducted experiments to quantitatively evaluate the obtained unbiased dental template by applying it to label teeth in the teeth segmentation images. First, we randomly selected 32 CBCT images from our digital archive, and segmented teeth from the CBCT images using the learning-based method. Then, we non-rigidly registered the labeled image of the unbiased dental template to each segmentation image, obtaining a warped labeled image. For each tooth in the segmentation image, we measured the Dice similarity coefficient (DSC) between the tooth and the teeth in the warped image, and assigned the label with the template tooth that had the largest DSC. As a comparison, we also performed automatic teeth labeling using the selected teeth template following the same method. According to the results, the mean success rate of the teeth labeling achieved by our unbiased dental template was 93.4%, which was higher than the success rate achieved by the selected teeth template (91.7%).

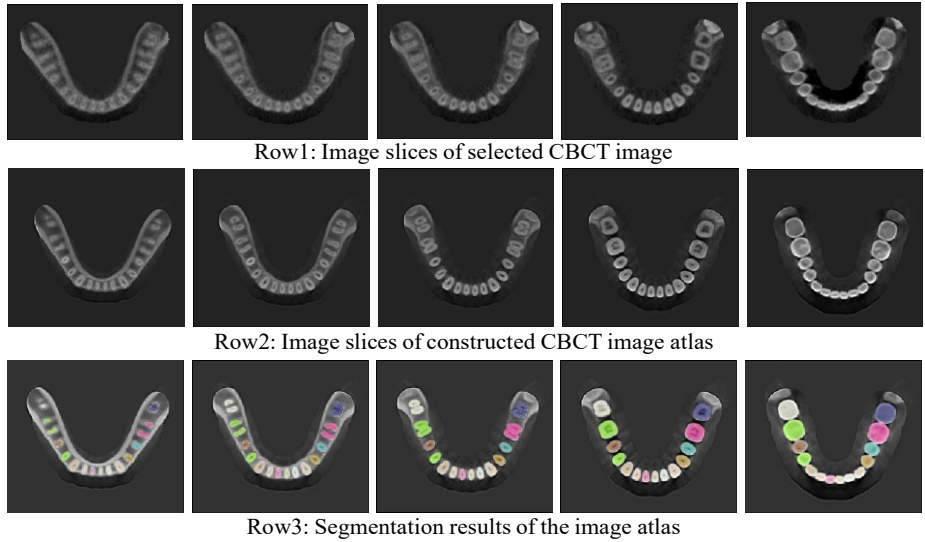

Figure 4: The dental atlas constructed by our proposed method. The first row presents the slice images of a selected CBCT image template. The second row shows slice images of the constructed CBCT image atlas, and the third row gives segmentation results of the slice images in the second row.

### 3.2. Parametric Dental Models

We excluded 21 subjects with missing teeth from the studied dataset when constructing parametric dental models. For each CBCT image of the remaining 138 subjects, we reconstructed a dental model from its segmentation image. We then performed spatial alignment and point-to-point correspondence estimation of the dentition models and the individual tooth models across the 138 subjects by leveraging the unbiased dental template. Finally, we employed PCA to determine shape variations of the dentition and individual tooth models as subspaces of their parametric models.

According to the results, the first 31 principal components (PCs) of the constructed parametric dentition model explained more than 85% of the shape variations of dentition models. The first three PCs, which explained about 45.0% of the whole variations, are shown in Fig. 6. Specifically, the first PC (i.e., PC1) exhibited a variation from underbite teeth to overjet teeth, with an increase in dental arch length and a decrease in dental arch width. In PC2, there was a decrease in teeth length and a slight increase in overjet from the left dentition to the right dentition. PC3 showed a decrease in labiolingual/buccolingual inclination of the upper and lower teeth.

To construct the parametric model of each tooth, we aligned each tooth in the dental models to its corresponding tooth template and estimated point-to-point correspondence among them for PCA analysis. Fig. 7 shows the first three PCs of shape variations of seven different teeth, i.e., from tooth11 to tooth17. For all seven teeth, their first PCs demonstrated an increase in tooth size. For the incisors (tooth11 and tooth12), canine (tooth13), and premolars (tooth14 and tooth15), their second PCs represented an increase in tooth width, including tooth crown and root, while their third PCs indicated an increase in tooth root with a slight decrease in tooth crown width. For the molars (tooth16 and tooth17), their second PCs described an increase in tooth width with a slight decrease in tooth length. PC3 of tooth16 indicated an extension of one tooth root, while PC3 of tooth17 represented an increase in tooth roots with a slight decrease in tooth crown width.

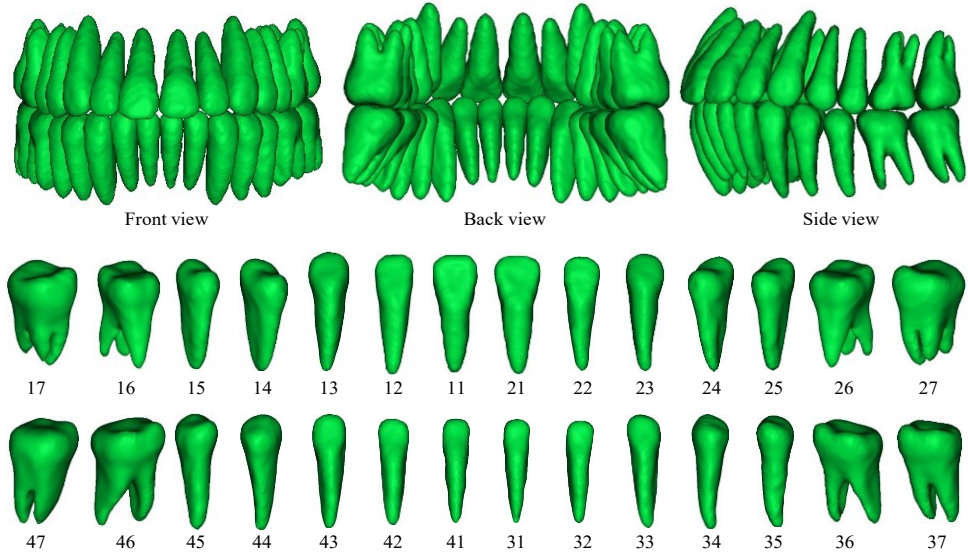

Figure 5: The unbiased dental template generated from the constructed dental atlas. The first row shows the different views (i.e., front view, back view, and side view) of the dental template reconstructed from the dental atlas. The second and third rows show the templates of different teeth.

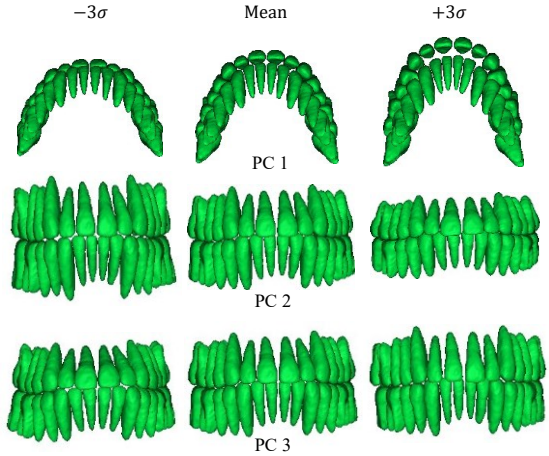

Figure 6: The first three PCs of the constructed parametric dentition model.

## 4. Discussions and Conclusions

Our paper presents an accurate dental atlas construction framework to generate an unbiased dental template. The evaluation results show that the constructed unbiased dental template outperforms the selected dental template, which has subjective bias. This indicates that our unbiased dental template has more typical shapes and location information of teeth and higher cross-individual validity than the biased template, thus facilitate image processing and analysis in digital dentistry. Besides teeth labeling in the evaluation, the unbiased dental template can also be applied to various digital dentistry applications, such as spatial alignment of teeth models (images) and correspondence estimation across different teeth models.

By leveraging the unbiased dental template, we can accurately construct parametric models for the dentition and individual tooth from the studied dataset. From the constructed parametric dental models, we can easily observe the teeth shape variations and their proportions in the subspace of the parametric dental models. The parametric dental models have various applications in digital dentistry. Specifically, the parametric dental models can be used as shape constraints in teeth segmentation and teeth deformable registration. The parametric models can also be applied to 3D teeth reconstruction from 2D pho-tos or 2D dental x-ray images. Other potential applications of the parametric dental models include teeth completion, normal teeth prediction in orthodontic planning, and more.

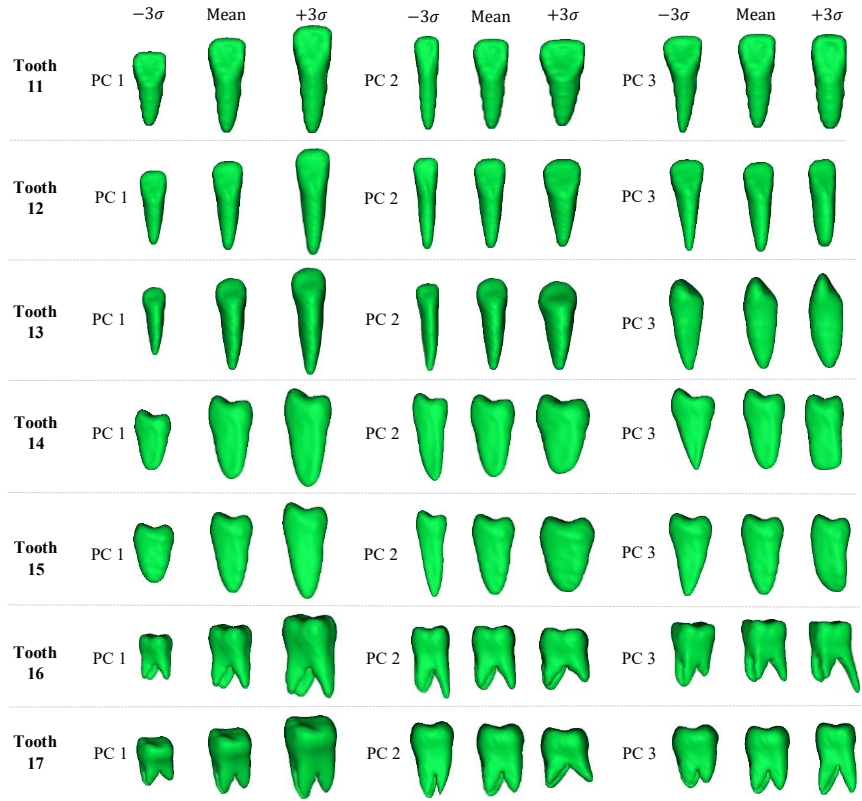

Figure 7: The first three PCs of the parametric models of seven different teeth, i.e., tooth 11 to tooth 17.

In this study, we present the results of unbiased templates and parametric models constructed for the dentition and the individual teeth. Based on these results, one can easily obtain unbiased templates and parametric models of upper teeth rows, lower teeth rows or other teeth clusters. This can extend the applications of the constructed unbiased dental template and parametric dental models to new scenarios, where the upper or lower teeth row is focused. In the future, we plan to further evaluate the performance of the unbiased dental template and parametric dental models by applying them to more specific applications.

## Acknowledgments

This work was supported in part by National Natural Science Foundation of China (grant number 62131015), Science and Technology Commission of Shanghai Municipality (STCSM) (grant number 21010502600), and The Key RD Program of Guangdong Province, China (grant number 2021B0101420006).